\RequirePackage{fix-cm}
\documentclass[onecolumn,10pt]{svjour3}
\smartqed  
\usepackage{graphicx}
\usepackage{fixltx2e}
\usepackage[misc]{ifsym}
\usepackage{float}
\usepackage[multi-part-units=single]{siunitx}
\usepackage[font=small,labelfont=bf,tableposition=top]{caption}
\usepackage{mathtools}
\usepackage{xcolor}
\usepackage{amsmath}
\usepackage{amssymb}
\usepackage{multirow}

\DeclareCaptionLabelFormat{andtable}{#1~#2  \&  \tablename~\thetable}
\begin{document}

\title{Deep Transfer Learning Methods for Colon Cancer Classification in Confocal Laser Microscopy Images}


\titlerunning{Deep Transfer Learning Methods for Colon Cancer Classification with CLM}        

\author{Nils Gessert$^{1}$$^{*}$, Marcel Bengs$^{1}$$^{*}$, Lukas Wittig$^{2}$, Daniel Dr\"omann$^{2}$, Tobias Keck$^3$, Alexander Schlaefer$^{1}$, David B. Ellebrecht$^{3}$}

\authorrunning{Gessert et al.} 

\institute{\Letter \quad Nils Gessert, \email{nils.gessert@tuhh.de}, Tel.: +49 (0)40 42878 3389, https://orcid.org/0000-0001-6325-5092 \\ $^{*}$ Authors contributed equally  \\ $^1$ Institute of Medical Technology, Hamburg University of Technology, Hamburg, Germany \\ $^2$ Department of Pulmology, University Medical Centre Schleswig-Holstein, L\"ubeck, Germany \\ $^3$ Department of Surgery, University Medical Centre Schleswig-Holstein, L\"ubeck, Germany}

\date{Preprint. Accepted for publication in IJCARS.}

\maketitle

\begin{abstract}

\textit{Purpose}

The gold standard for colorectal cancer metastases detection in the peritoneum is histological evaluation of a removed tissue sample. For feedback during interventions, real-time in-vivo imaging with confocal laser microscopy has been proposed for differentiation of benign and malignant tissue by manual expert evaluation. Automatic image classification could improve the surgical workflow further by providing immediate feedback.  

\textit{Methods} 

We analyze the feasibility of classifying tissue from confocal laser microscopy in the colon and peritoneum. For this purpose we adopt both classical and state-of-the-art convolutional neural networks to directly learn from the images. As the available dataset is small, we investigate several transfer learning strategies including partial freezing variants and full fine-tuning. We address the distinction of different tissue types, as well as benign and malignant tissue. 

\textit{Results} 

We present a thorough analysis of transfer learning strategies for colorectal cancer with confocal laser microscopy. In the peritonuem, metastases are classified with an AUC of $97.1$ and in the colon the primarius is classified with an AUC of $73.1$. In general, transfer learning substantially improves performance over training from scratch. We find that the optimal transfer learning strategy differs for models and classification tasks. 

\textit{Conclusions} 

We demonstrate that convolutional neural networks and transfer learning can be used to identify cancer tissue with confocal laser microscopy. We show that there is no generally optimal transfer learning strategy and model as well as task-specific engineering is required. Given the high performance for the peritoneum, even with a small dataset, application for intraoperative decision support could be feasible.

\end{abstract}

\section{Introduction} \label{intro}

Colorectal cancer is very common and it is often associated with metastatic spread \cite{torre2015global}. In particular, peritoneal carcinomatosis (PC) can arise in later stages of development which often shortens patient survival times substantially \cite{verwaal2005long,franko2012treatment}. Thus, early and reliable detection of metastases is crucial. Diagnosis with typical external imaging techniques such as computed tomography (CT) and magnetic resonance imaging (MRI) is difficult for PC as a very high resolution is required. For example, preoperative CT has been shown to be ineffective to detect individual peritoneal tumor deposits and the interobserver variability among experts was significant \cite{de2004peritoneal}. Also, integrated PET/CT did not provide sufficient information for accurate assessment \cite{dromain2008staging}. For MRI, studies have shown improvement over assessment with CT only \cite{low2000extrahepatic,iafrate2012peritoneal} but overall, its resolution is still a limitation \cite{gonzalez2009imaging}.  Therefore, exploratory laparoscopy is generally employed to investigate the presence of PC \cite{ishigami2014clinical}. 

Recently, a new intraoperative device using confocal laser microscopy (CLM) has been introduced which provides submicrometer image resolution \cite{ellebrecht2018confocal}. In the study, ten rats received colon carcionoma cell implants in the colon and peritoneum. After a growth period, laparotomy with in-vivo CLM was performed. CLM images of healthy and malignant colon tissue, as well as healthy and malignant peritoneum were acquired. It was shown that experts are able to distinguish different tissue types as well as healthy and malignant tissue from CLM. This raises the question whether image processing techniques can be used to automatically classify different tissue types. This could enable faster and improved intraoperative decision support with CLM. 

Recently, automatic tissue characterization has been successfully addressed using deep learning methods such as convolutional neural networks (CNNs) for semantic segmentation and classification \cite{Litjens.2017,goceri2017deep}. For example, skin cancer classification at dermatologist-level performance was achieved \cite{esteva2017dermatologist}. However, the datasets for this and related studies are large and commonly, datasets for medical learning tasks are small \cite{shen2017deep}. This can be problematic as insufficient data for optimal training might lead to overfitting and limited generalization. This is particularly important for deep learning models which can be prone to overfitting due to their large number of trainable parameters. To overcome this issue, transfer learning methods have been proposed where a deep learning model is first pretrained on a different, large dataset \cite{bengio2012deep}. Then, information from the source domain can be transferred to the (medical) target domain using strategies such as "off-the-shelf" features, partial layer freezing, or full fine-tuning \cite{hoo2016deep}. While this has been successfully applied for medical learning tasks \cite{gessert2018automatic}, there is no single solution for all problems and the optimal transfer learning strategy is highly dependent on the imaging modality and dataset size \cite{tajbakhsh2016convolutional}.

Automatic analysis of CLM images has been proposed for different tissue types such as human skin \cite{rajadhyaksha1995vivo}, the cornea \cite{niederer2007age} or the oral cavity \cite{aubreville2017automatic}. Recently, deep learning methods have been applied to CLM and similar modalities. For example, CNNs have been used for oral squamous cell carcinoma classification \cite{aubreville2017automatic} and motion correction with CLM \cite{aubreville2018deep}. Similarly, skin images from CLM have been used with CNN-based classification \cite{wiltgen2016automatic}. For the gastrointestinal tract, CNNs have been used to distinguish three classes of Barret's esophagus \cite{hong2017convolutional}. Also, brain tumor classification with CNNs and CLM has shown promising results \cite{Izadyyazdanabadi.2018}. For example, a CNN has been used to differentiate CLM images with and without diagnostic value for a physician during surgery \cite{izadyyazdanabadi2017improving}. Also, weakly-supervised localization has been used to derive local information in CLM images from image-level labels only \cite{izadyyazdanabadi2018weakly}.

So far, deep learning-based classification of colorectal cancer from CLM images has not been addressed. Also, while several approaches have used CLM and CNNs for other problems \cite{izadyyazdanabadi2018convolutional}, there is no analysis of transfer learning properties for colorectal cancer with CLM. Therefore, we study deep learning-based colon cancer classification from CLM images with a variety of transfer learning methods from the ImageNet dataset. We consider training from scratch, partial layer freezing, "off-the-shelf" features and full fine-tuning to investigate how transferable ImageNet features are to CLM. We perform this study with the classic models VGG-16 \cite{simonyan2014very} and Inception-V3 \cite{szegedy2016rethinking} as well as the state-of-the-art architectures Densenet \cite{Huang.2016} and squeeze-and-excitation networks \cite{hu2018squeeze} to analyze the consistency of transfer strategies across architectures. We consider the classes healthy colon (HC), malignant colon (MC), healthy peritoneum (HP) and malignant peritoneum (MP). Based on these classes, we address three binary classification tasks with CLM. First, we consider the differentiation of organs (HP vs. HC). Then, we study the detection of malignant tissue in two types of organs (HP vs. MP and HC vs. MC). This allows us to study variations across different classification tasks for CLM.
A preliminary version of this paper was presented at the BVM Workshop 2019 \cite{gessert2019colon}. We substantially revised the paper, extended the review of the literature and performed more experiments with additional transfer strategies and more architectures.
This paper is structured as follows. First, we describe our models and transfer learning stratgies and the data set we use in Section~\ref{sec:methods}. Then we report our results in Section~\ref{sec:results} and discuss them in Section~\ref{sec:discussion}. Last, we conclude in Section~\ref{sec:conclusion}.

\section{Methods} \label{sec:methods}

\subsection{Model Architectures and Training Strategies} \label{sec:models}

\begin{figure}
\centering
\includegraphics{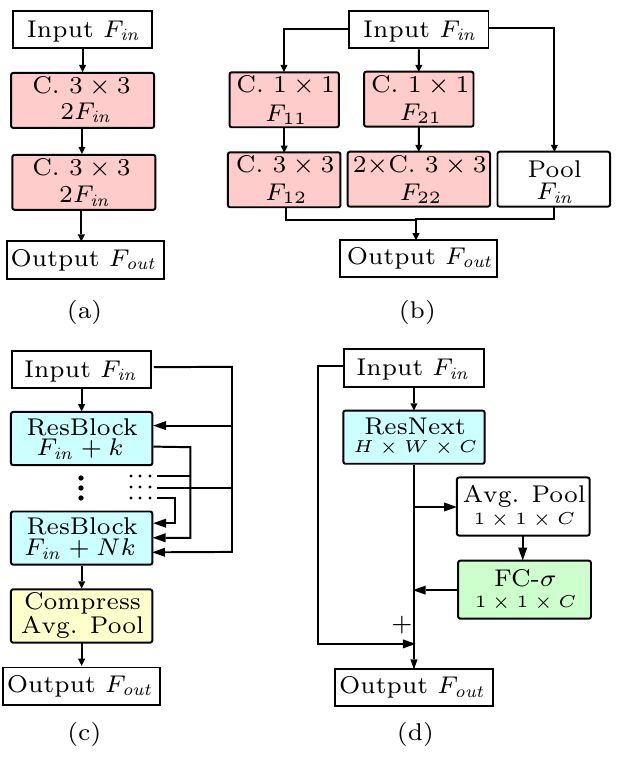} 
\caption{The building blocks of the models we use. The building blocks from CNN architectures as indicated in Figure~\ref{fig:transfer}. We employ VGG-16 (a), Inception-V3 (b), Densenet121 (c) and SE-Resnext50 (d). $F$ denotes the number of feature maps in each block. The Conv blocks also contain ReLU activations and batch normalization for VGG-16 (a). SE-Resnext50 is shown in simplified form without its bottleneck in the SE module. FC-$\sigma$ is a fully-connected layer with sigmoid activation. C. is an abbreviation for convolutional layers. Note that Inception-V3 employs multiple block variants and we show one example.}
\label{fig:models}
\end{figure}

First, we consider the classic model VGG-16 \cite{simonyan2014very} with the addition of batch normalization which enables faster training of the architecture by reducing the internal covariate shift \cite{Ioffe.2015}. The model itself is simple as it consists of several stacked convolutional layers without further augmentation. In between blocks of two to three convolutional layers with kernel sizes of $3\times 3$ and $1\times 1$, max pooling reduces the spatial dimensions. Subsequent convolutions double the number of feature maps. A building block of the architecture is shown in Figure~\ref{fig:models} (top left). Due to its simple structure, the architecture can serve as a baseline. 

Second, we employ Inception-V3 \cite{szegedy2016rethinking}. The model consists of multiple Inception blocks which follow two core design principles. First, the blocks have a multi-path structure, i.e., the input feature maps are processed in parallel by different convolution and pooling operations. At the block's output, the feature maps from all paths are concatenated. Second, the convolutional paths perform a reduction operation that downsizes the feature map dimension with $1 \times 1$ kernels. Then, computationally more expensive $3\times 3$ convolutions process the lower dimensional representations. The output feature map size is increased if the spatial dimensions are reduced inside the block which avoids representational limitations.


The idea of reduction and expansion has also found its way into the Resnet architecture \cite{He.2016} which is a core component of the next two models. Resnets learn a residual instead of a full feature transformation by using skip connections. In detail, a Resnet block (ResBlock) computes

\begin{equation}
	x_{(l)} = a(\mathcal{F}(x_{(l-1)},\theta_{(l)}) + x_{(l-1)})
\end{equation}

where $x_{(l)}$ is the block output, $x_{(l-1)}$ is the block input, $a$ is a ReLU activation \cite{nair2010rectified} and $\mathcal{F}$ represents two convolutional layers with parameters $\theta_{(l)}$. The skip connection enables better gradient propagation for improved training. 

Third, we consider Densenet121 \cite{Huang.2016}, a state-of-the-art architecture which strives for more efficiency by introducing extensive feature reuse. In particular, within one DenseBlock, features computed in previous layers are also fed into the subsequent layers. To keep the feature map sizes moderate, compression blocks reduce the feature maps between DenseBlocks. The DenseBlock is shown in Figure~\ref{fig:models} (bottom left). 

Fourth, we adopt the architecture SE-Resnext50 \cite{hu2018squeeze}. At its core, the model uses Resnext blocks \cite{Xie.2017} which are an extension of Resnet. Here, the single convolutional path $\mathcal{F}$ is split into multiple paths with individual layers which increases representational power. The key addition in SE-Resnext50 is the use of squeeze-and-excitation (SE) modules which recalibrate the feature maps learned by Resnext blocks. These modules have shown improved performance with only a minimal increase in the number of parameters. The concept is shown in Figure~\ref{fig:models} (bottom right).  

\begin{figure}
\centering
\includegraphics{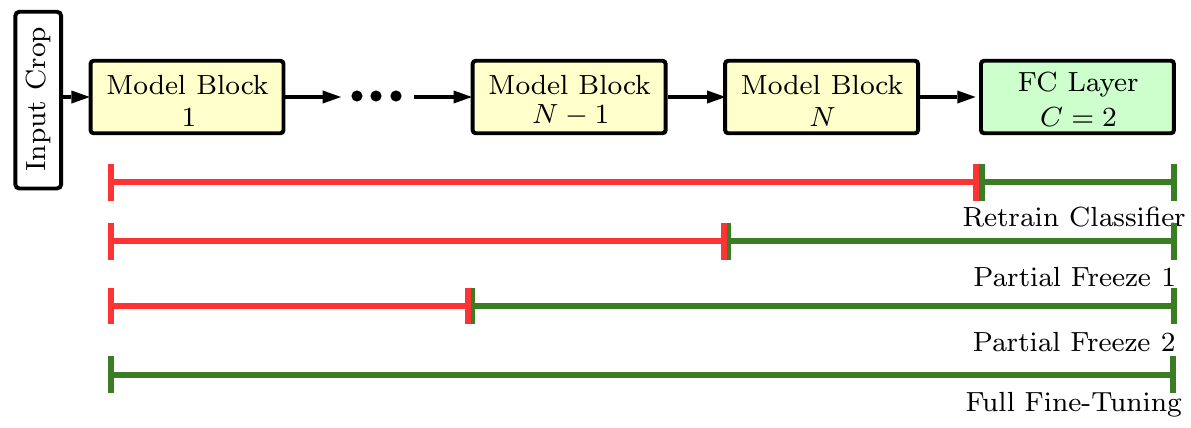}
\caption{The different transfer learning scenarios we investigate. Model Block refers to one of the blocks shown in Figure~\ref{fig:models}. Green indicates that blocks are retrained. Red indicates that blocks are frozen with their weights having been trained on ImageNet.}
\label{fig:transfer}
\end{figure}

Due to the small dataset size, we study several transfer learning strategies where the above-mentioned models are trained on ImageNet. We cut off the last layer of all models and replace it with a fully-connected layer with two outputs for binary classification. We apply a softmax layer on top and the final classification output is the class with the highest probability. We train a separate model for each of our binary classification tasks. 

As a baseline, we consider training from scratch, i.e. all weights are randomly initialized. Then, we use several different transfer learning strategies illustrated in Figure~\ref{fig:transfer}. The first transfer approach follows the "off-the-shelf" features idea. Here, only the new classifier is retrained on features extracted by the pretrained CNN. We also consider two partial freezing methods, where an initial part of the network remains frozen and the part closer to the classifier is retrained. We chose the freezing points block-wise, i.e. we do not cut into building blocks. Last, we consider full fine-tuning where all weights in the network are retrained with a small learning rate. The different strategies represent different abstractions of feature transfer between ImageNet and CLM images.

To further improve generalization, we employ online data augmentation with random image flipping and random changes in brightness and contrast. Furthermore, we use random cropping with crops of size $224\times 224$ ($299\times 299$ for Inception-V3) taken from the full images of size $384\times 384$. We use the Adam algorithm for optimization. We adapt learning rates and the number of training epochs for the different transfer scenarios. We use a cross-entropy loss function with additional weighting to account for the slight class imbalance. In detail, we multiply the loss of a training example by $N/n_i$ where $N$ is the total number of training examples in the current fold and $n_i$ is the number of examples belonging to class $i$ in the current fold. In this way, underrepresented classes receive a higher weighting in the loss function. During evaluation, we use mutli-crop evaluation with $N_c = 36$ evenly spread crops over the images. This ensures that all image regions are covered with large overlaps between crops. The final predictions are averaged over the $N_c$ crops. We implement our models in PyTorch.

\subsection{Dataset and Experiments}

\begin{figure}
\centering
\includegraphics[width=0.24\textwidth]{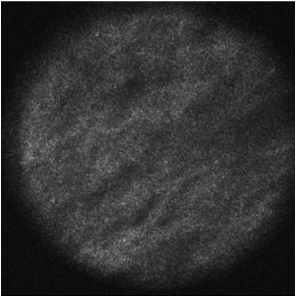}
\includegraphics[width=0.24\textwidth]{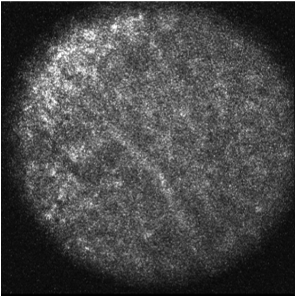}
\includegraphics[width=0.24\textwidth]{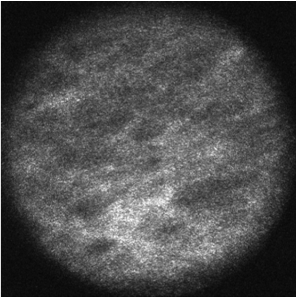}
\includegraphics[width=0.24\textwidth]{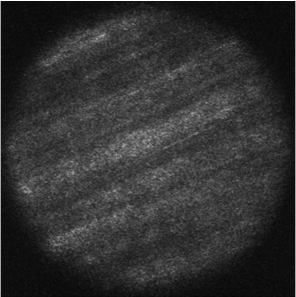}
\caption{Examples for the different classes. Malignant colon tissue, healthy colon tissue, malignant peritoneum tissue and healthy peritoneum tissue are shown from left to right.}
\label{fig:examples}
\end{figure}

The dataset was collected in a previous study conducted at the University Hospital Schleswig-Holstein in L\"ubeck where expert assessment of CLM images in the colon area was evaluated \cite{ellebrecht2018confocal}. A custom intraoperative device with integrated CLM (Karl Storz GmbH \& Co KG, Tuttlingen, Germany) was built. The image resolution was $384\times 384$ pixels which covers a field of view of  $\SI{300}{\micro\metre} \times \SI{300}{\micro\metre}$. In the study, ten rats received colon adenocarcinoma cell implantation in the colon and peritoneum with a growth time of seven days. Then, laparotomy was conducted and images of healthy colon tissue, malignant colon tissue, healthy peritoneum tissue and malignant peritoneum tissue were obtained. Example CLM images for each tissue type are shown in Figure~\ref{fig:examples}. After removal of low quality images, 1577 images remained with 533 belonging to class HC, 309 belonging to class MC, 343 belonging to class HP and 392 belonging to class MP. Note that some subjects are missing classes such that, on average, six subjects per class remain. Ground-truth annotation of all images was obtained by tissue removal of the scanned areas and subsequent histological evaluation.

Due to the small dataset size, we chose a cross-validation scheme where images from one subject are left for evaluation and training is performed on the remaining ones. Thus, all reported results are the mean value of, on average, six training scenarios with six different folds. Based on the four classes, we address three binary classification problems. First, we consider HC vs. HP, i.e., we investigate the feasibility of distinguishing the different organs in CLM. Then, we consider the differentiation of healthy and malignant tissue with the two binary classification problems HP vs. MP and HC vs. MC. We report the accuracy, sensitivity, specificity, F1-score and AUC. We use the AUC as the main metric as it is threshold-independent.

\section{Results} \label{sec:results}

\begin{figure}
\centering
\includegraphics{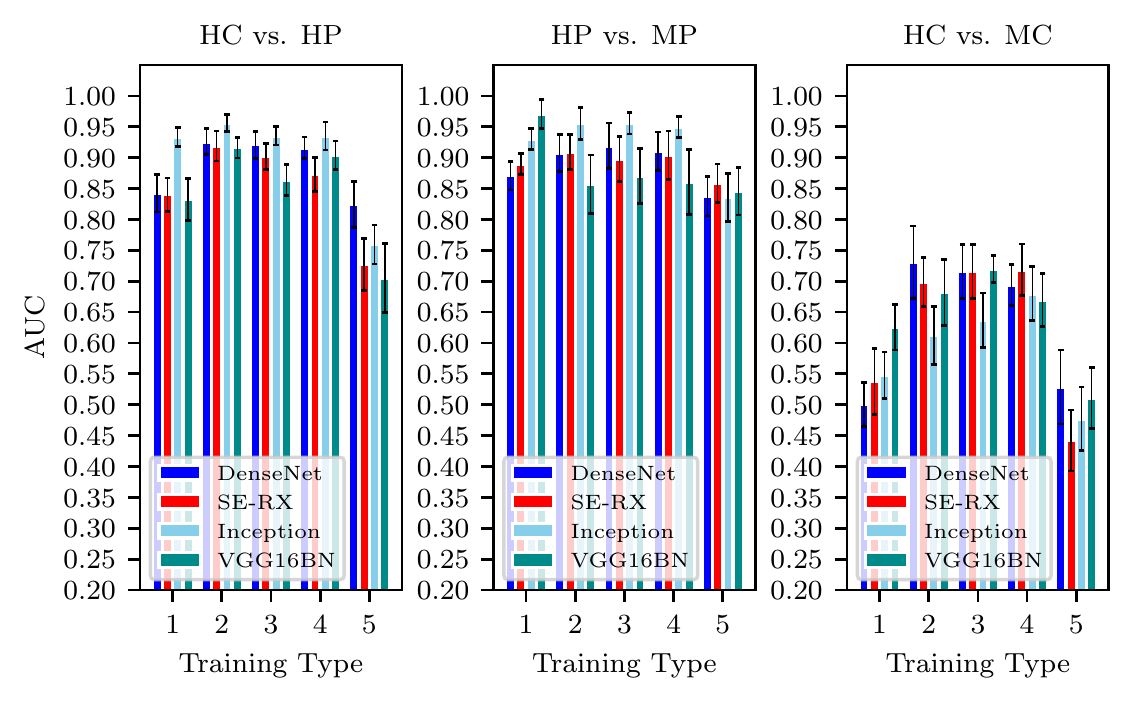}
\caption{AUC values of all applied architectures for the different classification problems. We evaluate the following training types: (1) retrain classifier, (2) partial freeze 1, (3) partial freeze 2, (4) full fine-tuning, (5) training from scratch. For each value the standard deviation over multiple folds is represented by an error bar.}
\label{fig:AUC}
\end{figure}

First, we compare the different transfer learning scenarios described in Section~\ref{sec:methods} across all architectures for each classification scenario, see Figure~\ref{fig:AUC}. In general, the AUC is very high for the differentiation of different healthy tissue types and healthy and malignant peritoneum tissue. The AUC for classifying malignant colon tissue is substantially lower. Also, the standard deviation is higher for this task. Training from scratch performs worst for all architectures and classification scenarios.

Regarding the transfer learning scenarios, training from scratch performs worst for all classification scenarios. For two of the three scenarios, only retraining the classifier shows substantially lower performance than other transfer scenarios. There are no clear trends between the partial freezing and fine-tuning scenarios. 

\begin{figure}
\centering
\includegraphics{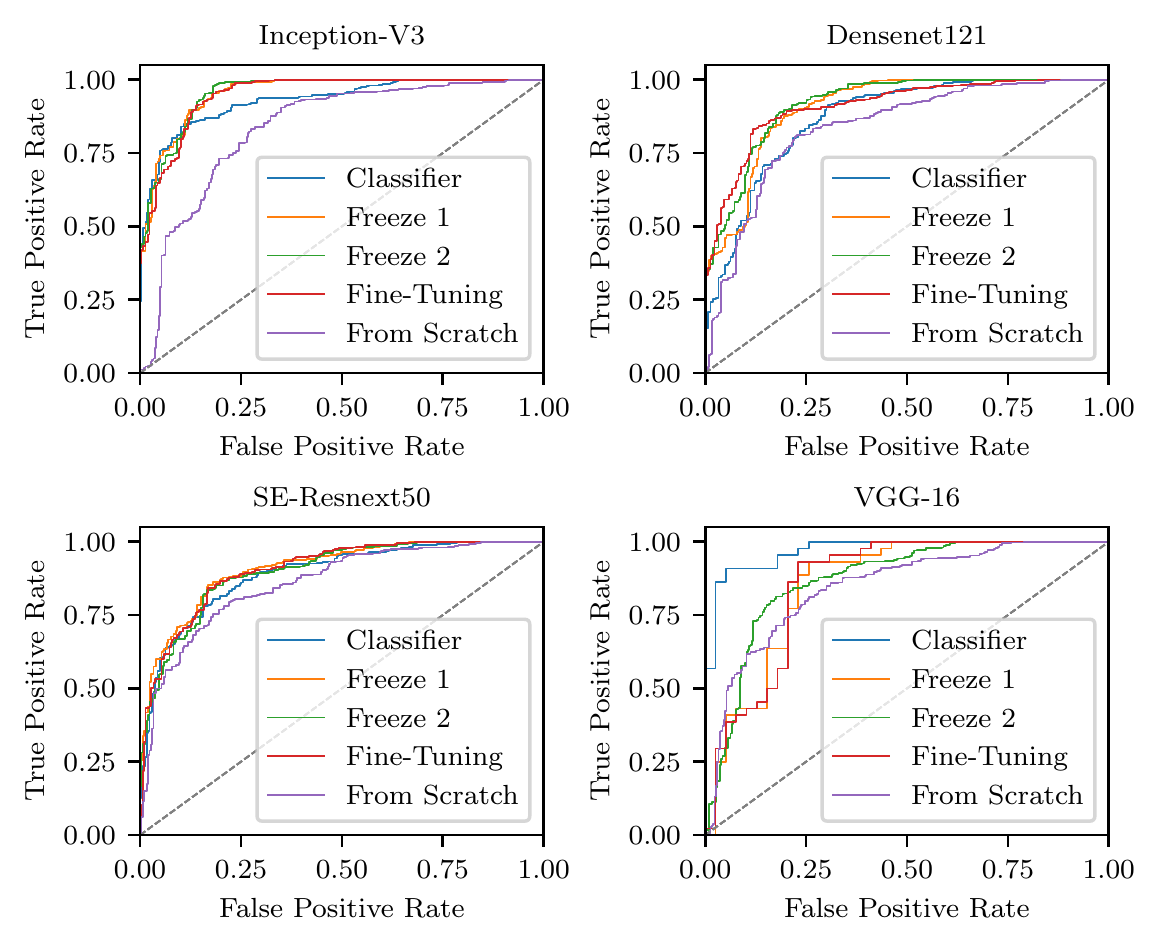}
\caption{ROC curve for the different architectures and the different training types, shown for the classification of HP vs. MP.}
\label{fig:ROC}
\end{figure}

Second, we go into more details for the classification task HP vs. MP. Figure~\ref{fig:ROC} shows the ROC curves for all models with all transfer learning scenarios for the classification task. Operating points with a good trade-off in the upper left corner vary for each model. For VGG-16, retraining the classifier only stands out. For Densenet121, partial freezing performs well. For Inception-V3 and SE-Resnext50, partial freezing and fine-tuning perform similar.

\renewcommand{\arraystretch}{1.3}
\begin{table}[t]
\caption{The best performing transfer learning method for each model and classification task. Densenet refers to the Densenet121 model, SE-RX50 refers to the SE-Resnext50 model. For each training scenario, the best performing configuration is marked bold. All values are given in percent. The sensitivity is given with respect to the cancer class and for the case of organ differentiation it is given with respect to the peritoneum class.}
\label{tab:result}
\begin{tabular}{llllllll}
\hline 
 &  & Type & Accuracy & Sensitivity & Specificity & F1-Score & AUC\tabularnewline
\hline 
\parbox[t]{2mm}{\multirow{4}{*}{\rotatebox[origin=c]{90}{HC vs. HP}}} & Inception & Freeze 1 & 87.7 & 79.9 & 94.4 & 90.4 & \textbf{95.7}\tabularnewline
 & Densenet & Freeze 1 & \textbf{91.2} & \textbf{82.8} & 95.3 & \textbf{91.9} & 92.6\tabularnewline
 & SE-RX50 & Freeze 1 & 85.8 & 78.5 & \textbf{96.3} & 91.3 & 91.9\tabularnewline
 & VGG-16 & Freeze 1 & 82.5 & 74.9 & 91.8 & 87.2 & 91.6\tabularnewline
\hline 
\parbox[t]{2mm}{\multirow{4}{*}{\rotatebox[origin=c]{90}{HP vs. MP}}} & Inception & Freeze 2 & 85.9 & 86.6 & \textbf{87.0} & 86.8 & 95.6\tabularnewline
 & Densenet & Freeze 2 & 83.3 & 84.6 & 83.2 & 84.0 & 91.9\tabularnewline
 & SE-RX50 & Freeze 1 & 81.7 & 84.6 & 83.2 & 84.0 & 90.9\tabularnewline
 & VGG-16 & Classifier & \textbf{88.0} & \textbf{91.0} & 84.6 & \textbf{87.9} & \textbf{97.1}\tabularnewline
\hline 
\parbox[t]{2mm}{\multirow{4}{*}{\rotatebox[origin=c]{90}{HC vs. MC}}} & Inception & Fine-Tuning & 63.1 & 71.0 & 57.0 & 63.7 & 68.0\tabularnewline
 & Densenet & Freeze 1 & \textbf{70.0} & \textbf{72.9} & 64.1 & \textbf{69.1} & \textbf{73.1}\tabularnewline
 & SE-RX50 & Fine-Tuning & 63.7 & 66.7 & \textbf{65.9} & \textbf{69.1} & 71.8\tabularnewline
 & VGG-16 & Freeze 2 & 63.5 & 67.6 & 64.2 & 68.1 & 72.0\tabularnewline
\hline 
\end{tabular}
\end{table}

Third, an overview of the best performing transfer strategies is shown in Table~\ref{tab:result}. Comparing individual results for each architecture, no model clearly outperforms the others consistently. In general, Densenet121 performs slightly better across the tasks. The optimal transfer strategy differs across models and classification tasks. For HC vs. HP and for Densenet121 in general, the partial freezing method performs best. 

\begin{figure}
\centering
\includegraphics{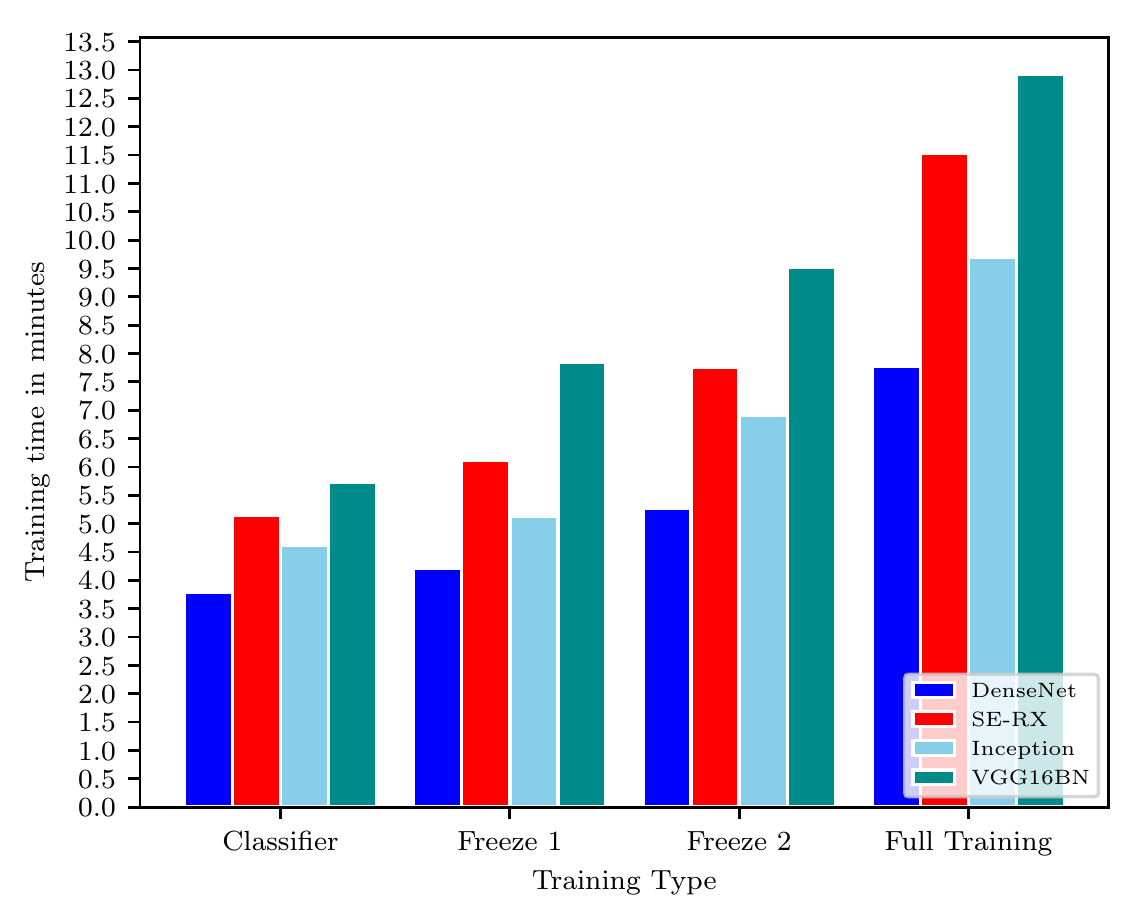}
\caption{Training times for 90 epochs of all applied architectures for the different training scenarios for the classification task HP vs HC. Note that for training from scratch the same number of parameters is trained as for full-fine tuning. Thus, training times are equivalent for the two cases.}
\label{fig:Training_Times}
\end{figure}

Last, we provide training times for all architectures and training scenarios, see Figure~\ref{fig:Training_Times}. In general, freezing more weights during training reduces the overall training time. Furthermore, training time loosely scales with the number of trainable parameters as VGG-16 contains the most parameters and shows the longest training times, followed by SE-Resnext50.


\section{Discussion} \label{sec:discussion}

We study deep transfer learning methods for CLM images for three binary classification problems. Automatic decision support with CLM during interventions could improve workflow with immediate feedback on the tissue properties. For this purpose we investigate the use of CNNs with four different architectures and five training scenarios.

\textbf{The three classification tasks.} As a baseline, differentiating healthy colon and peritoneum tissue works well with an AUC over $0.90$ for partial freezing across all models, see Figure~\ref{fig:AUC}. This indicates that discriminative features for different organs can be learned from CLM images. Similarly, for classification of metastases in the peritoneum the AUC is around $0.90$ for all transfer learning scenarios. However, classifying healthy and malignant colon tissue performs substantially worse with an AUC of $\approx 0.70$ for partial freezing and fine-tuning. The task appears to be more difficult which is also reflected in a slightly higher standard deviation. This indicates higher uncertainty of model predictions. This could be caused by the heterogeneous appearance of colon tissue in different parts of the colon which complicates the learning task in conjunction with the small dataset size. Furthermore, during development, colon carcinoma cells transform from a healthy stage to adenoma and then carcinoma. At earlier stages, healthy and malignant cells can still have similar appearance which complicates the learning task. 

\textbf{Transfer learning scenarios.} Figure~\ref{fig:AUC} also provides an overview of the transfer strategies across all models. Clearly, transfer learning substantially outperforms training from scratch across all classification tasks which supports the effectiveness of transfer learning for medical image classification problems \cite{Shin.2016}. The results indicate that meaningful feature transfer from the natural image domain to CLM images is possible, although the images have a vastly different appearance. However, comparing transfer strategies, only retraining the classifier performs worse than other scenarios in two out of three classification tasks. This agrees with results of a previous study on transfer learning with CLM images in neurosurgery \cite{izadyyazdanabadi2018convolutional}. Here, the authors found that full fine-tuning outperforms retraining of the classifier only. However, in our case, retraining the classifier only also shows a high performance for the task HP vs. MP. This could be caused by fragile co-adaptation of weights \cite{yosinski2014transferable} which leads to large performance differences between the different classification tasks. For some tasks (e.g. HP vs. MP) recovery and reuse of potentially co-adapted weights might be feasible while reuse is impaired for other tasks (e.g. HC vs. MC). The partial freezing and fine-tuning strategies appear to be more consistent across tasks, however, the optimal strategy still differs. Overall, our results indicate that the transferability of features not only depends on the imaging modality but also the classification task. This adds to previous insights on transfer learning in the medical domain where the optimal transfer strategy was found to be modality and dataset size dependent \cite{tajbakhsh2016convolutional}. Comparing the partial freezing and fine-tuning strategies, performance is very close and there is no optimal strategy for each of the tasks. However, training times are also an aspect to consider for the different transfer learning strategies. As shown in Figure~\ref{fig:Training_Times}, freezing more parameters inside the architecture leads to reduced training times. Thus, partial freezing can be generally seen as advantageous as it often achieves similar performance as full tine-tuning while requiring less training time. For application, this insight could be useful when adopting and retraining models for cancer classification in other organs or when newer architectures are introduced.

\textbf{Different architectures for CLM.} To analyze the different transfer strategies further, we consider the ROC curves of each architecture for the HP vs. MP task, see Figure~\ref{fig:ROC}. For this task, using "off-the-shelf" features and only retraining the classifier performed considerably better than for the other tasks. As discussed before, this indicates that transfer learning scenarios are classification task-dependent. In detail, the ROC curves reveal that VGG-16 stands out in particular where retraining the classifier only performs best out of all transfer strategies. In transfer learning research, VGG-16 is still a popular general purpose feature extractor for numerous tasks \cite{herath2017going,Litjens.2017}. For the other architectures, the optimal strategy differs. For example, for Densenet121, the partial freezing methods show good operating points in the upper, left corner of the ROC-curve. For Inception-V3 and SE-Resnext50, partial freezing and fine-tuning perform similar with no clearly superior method. This indicates that the choice of transfer learning method depends on the architecture. This should be expected, as the models have very different block types and each freezing type fixes a different number of parameters. The detailed results in Table~\ref{tab:result} with additional metrics underline this insight. There is no optimal transfer learning strategy and the best performing strategy varies for different architectures and classification tasks. Overall, we demonstrate that transfer learning has an impact on performance, however, there is no simple rule-of-thumb for optimal transfer learning with CLM. Our results show that examining different freezing strategies can considerably improve performance for individual models.

\section{Conclusion} \label{sec:conclusion}

We investigate the feasibility of colon cancer classification in CLM images using CNNs and multiple transfer learning scenarios. Using in-vivo images of healthy and malignant colon and peritoneum tissue obtained from ten subjects, we adopt four architectures and five transfer learning scenarios for three classification problems with CLM. Our results show that different organs as well as healthy and malignant peritoneum tissue can be classified with deep transfer learning. We show that transfer learning from ImageNet is successful with CLM but the transferability of features is limited. We find that there is no single optimal model or transfer strategy for all CLM classification problems and that task-specific engineering is likely required for application. 
For future work, our results could be extended to more classification problems with CLM.

\section*{Compliance with Ethical Standards}

\small \textbf{Funding:} The authors have no funding to declare.

\small \textbf{Conflict of Interest:} 
The authors declare that they have no conflict of interest.

\small \textbf{Ethical Approval:} 
All procedures performed in studies involving animals were in accordance with the ethical standards of the institution or practice at which the studies were conducted.

\small \textbf{Informed Consent:} 
Informed consent was obtained from all individual participants included in the study.


\bibliographystyle{spmpsci} 
\bibliography{egbib}   

\end{document}